\documentclass[runningheads,a4paper]{llncs}

\usepackage{times}
\usepackage{latexsym}
\usepackage{url}

\usepackage{amsmath}
\usepackage{graphicx}
\usepackage{subfig}
\usepackage{color}
\usepackage{amsfonts,amssymb}
\usepackage{float}
\usepackage{bm} 

\title{CASICT Tibetan Word Segmentation System for MLWS2017}

\author{Jiawei Hu$^1$ \and Qun Liu$^{2,1}$
}
\authorrunning{CASICT Tibetan Word Segmentation System for MLWS2017}

\institute{$^1$Key Laboratory of Intelligent Information Processing, \\ Institute of Computing Technology, Chinese Academy of Sciences\\
        $^2$ADAPT Centre, School of Computing, Dublin City University\\
        {\tt \{hujiawei,liuqun\}@ict.ac.cn}}

\date{}

\begin{document}

\maketitle

\begin{abstract}
We participated in the MLWS 2017 on Tibetan word segmentation task, our system is trained in a unrestricted way, by introducing a baseline system and 76w tibetan segmented sentences of ours. In the system character sequence is processed by the baseline system into word sequence, then a subword unit (BPE algorithm \footnote{https://github.com/rsennrich/subword-nmt}) split rare words into subwords with its corresponding features,  after that a neural network classifier is adopted to token each subword into ``B,M,E,S'' label, in decoding step a simple rule is used to recover a final word sequence. The candidate system for submition is selected by evaluating the F-score in dev set pre-extracted from the 76w sentences. Experiment shows that this method can fix segmentation errors of baseline system and result in a significant performance gain. 
\end{abstract}

\section{Introduction}
We build the Tibetan word segmentation systems CASICT for MLWS 2017 Tibetan word segmentation task.
Our systems mainly consists of two parts in a pipeline form, a baseline system  (see section \ref{baseline}) as the preprocessor which segment the input  sequence of characters into words, and a refiner split the output words into subwords and token them by labels of ``B,M,E,S'', then a simple rule is used to  recover a segmented  word sequence. The preprocessor is pre-trained system for tibetan word segmentation on 1.3w segmented sentences,
We also introduce extra 76w segmented sentences to train the refiner, which is a deep neural network (DNN) for sequence labeling (see section \ref{Refiner}). 
Experiments show our Refiner is able to learn from baseline system and improve the performance on test set compared to the original system.

\section{System Description}
Our system consists of two main blocks, the first one is a classical system based on the work of \cite{sunmeng2014}, the second is one we call as ``Refiner'', is a neural network combined with subword preprocessing unit. We  describe them here in detail. 

\subsection{Baseline System}
\label{baseline}
Article \cite{sunmeng2014} proposed a discriminative model  with a word lattice and shortest—path based re-ranking strategy  for Tibetan word segmentation, we re-implement it as the baseline system and use a pre-trained model from \cite{sunmeng2014} without extra training procedure.

\begin{figure} 
    \centering
    \includegraphics[scale=0.5]{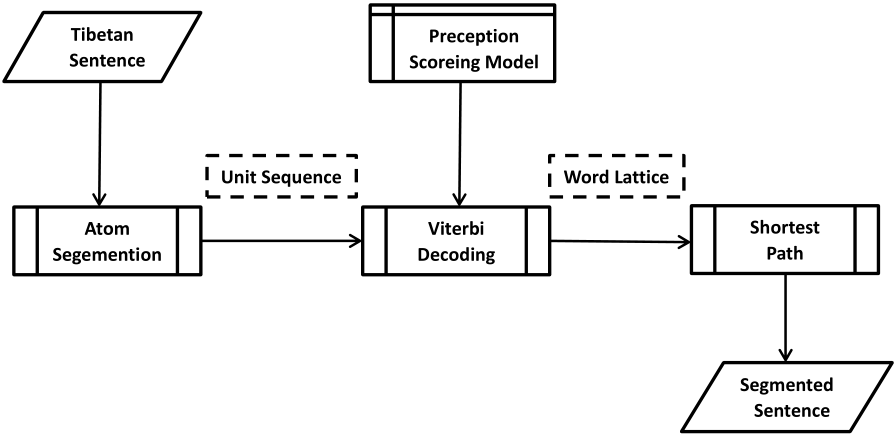}
    \caption{The architecture of baseline system}
    \label{fig:baseline}
\end{figure}

The baseline system pipeline is shown in Figure \ref{fig:baseline}, there are three key points as listed here:

\textbf{Atom segmentation:} it use syllable as word formation unit on the Tibetan segmentation to balance the segmentation precision on word level and performance on sentence level．

\textbf{Perception scoring model:} a simple scoring model is adopted to evaluate the candidate atom sequence which is segmented by tagging labels of 'b,m,e,s' on each atom, features is extracted by artificial rules in a window around each atom and then feed to a one layer liner perception to get the local tagging score.

\textbf{Shortest-path based re-ranking:} after the beam search procedure by viterbi decoding a word lattice is generated, then we weight each word span by rules (ie: scoring based on whether it`s a word in dictionary or not), and find the shortest path as the result segmented sequence.

Except for what are described above, a pre-segmentation strategy is applied to reduce the searching space by splitting sentence into sub-sequences by indication of special characters designed beforehand by expert.

\subsection{Deep Neural Network Refiner} \label{Refiner}
The Reiner is a neural network argumented with a subword preprocessing unit learned from tibetian corpus. The neural network we use here is a multi-layer recurrent ones unrolled by several bidirectional modified Long Short-Term Memory (LSTM) blocks \cite{Hochreiter1997}. To enhance message propagation in network, residual operation with gate is adopted to control information flow between layers horizontally. Here first give a brief introduction to the modified lstm block. 

\textbf{Modified lstm block:} a vanilla LSTM block with a gated residual connection, enable message flow efficiently in both time and spatial scale, which is described in equations (1) \ldots (5) bellow.

\begin{align}  
  \mathrm{i}_t &= \sigma( \bm{\mathrm{W}}_{xi}x_t + \bm{\mathrm{W}}_{hi}h_{t-1}) \\
  \mathrm{f}_t &= \sigma( \bm{\mathrm{W}}_{xf}x_t + \bm{\mathrm{W}}_{hf}h_{t-1}) \\
  \mathrm{c}_t &= \mathrm{f}_t \odot \mathrm{c}_{t-1} + \mathrm{i}_t \odot (\bm{\mathrm{W}}_{xc}x_t + \bm{\mathrm{W}}_{hc}h_{t-1}) \\ 
  \mathrm{o}_t &= \sigma( \bm{\mathrm{W}}_{xo}x_t + \bm{\mathrm{W}}_{ho}h_t) \\
  \mathrm{h}_t &= \mathrm{o}_t \odot \tanh \mathrm{c}_t + (1 -\mathrm{i}_t) \odot \mathrm{x}_t
\end{align}
Where $x_t$ is the input gate, $o_t$ is the output gate, $f_t$ is the forget gate, $c_t$ is the memory cell, $h_{t-t}$ is the hidden state of last time, and $h_t$ is the updated hidden state, we introducing a residual gate by reusing the input gate $i_t$, for $i_t$ is bounded in range (0,1), we set the residual gate as $1-i_t$, aim at balancing message updating in lstm cell and message gating operation.

\begin{figure} 
    \centering
    \includegraphics[scale=0.3]{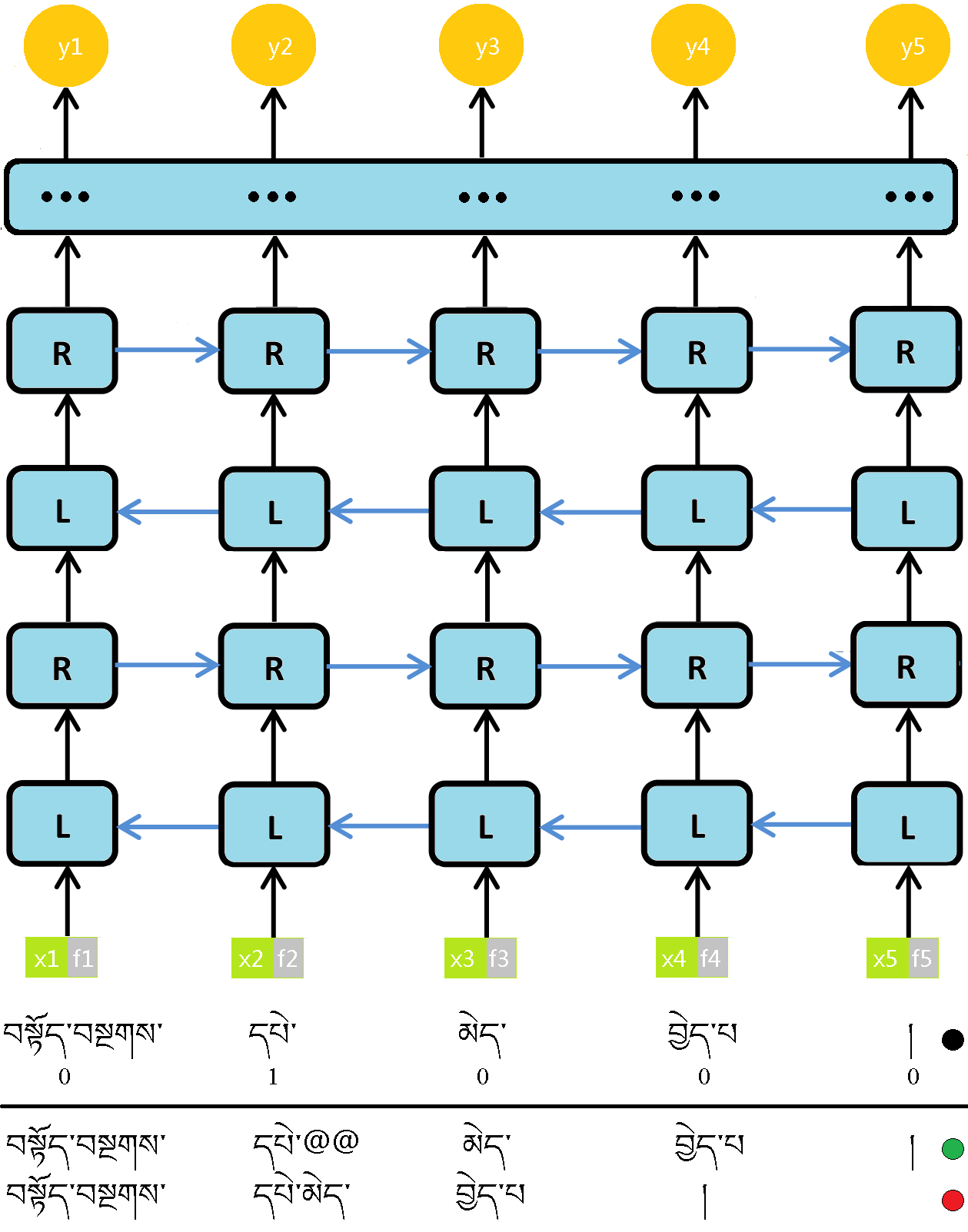}
    \caption{The architecture of Refiner system, sequence end with red dot is the output of baseline system, sequence end with blue dot is the output of Refiner`s subword preprocessing unit(SPU), sequences end with black dot are the input for neural network tagger.}
    \label{fig:refiner}
\end{figure}

We also unroll the modified LSTM unit in a stacked manner, by alternating direction between each layer(shown in Figure \ref{fig:refiner}), and get a deep neural network to encode input sequence to its representation, we stack 8 LSTM units in our system and share parameters between layers of same unrolling direction.

The Refiner`s pipeline is shown in Figure \ref{fig:refiner}, there are three key points as listed here:

\textbf{Subword preprocessing unit:} article \cite{Sennrich2015Neural} first introduced a subword segmentation method for neural machine translation based on
the byte pair encoding(BPE) compression algorithm, we use this method to split rare word into subwords with its corresponding feature, which indicates whether it`s a subword or not. Our BPE model is learned from 1664675 monolingual sentences clawed from tibetan websites. 

\textbf{Neural network sequence tagger:} we unroll the modified lstm as the way above to encode input subword sequence into its representations, each subword is represented by the hidden state of the lstm cell in last layer, and then feed into a liner softmax classify to get a label in 'B, M, E, S', where 'B' stands for the begining of a word, 'M' stands for the 'middle' of a word, 'E' means it`s the end of a word and 'S' means it`s a single word. In this step each subword is tagged with a particular label indicating its position in a word.

\begin{figure} 
    \centering
    \includegraphics[scale=0.3]{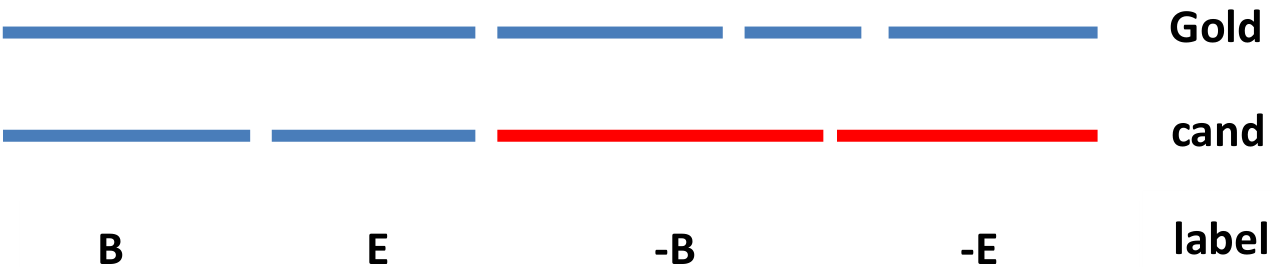}
    \caption{Example for subword sequence labeling}
    \label{fig:tagger}
\end{figure}

For the reason of the segmentation errors in baseline system, we design a argumented label set for the Refiner by adding labels '-B, -M, -E' to identify the wrongly segmented word span. For example in Figure \ref{fig:tagger}, ``Gold'' is the true segmentation sequence, ``cand'' is the subword sequence from the subword preprocessing unit, the first two tokens form a word in ``Gold'' but the last two are mis-segmented by baseline and cannot form any word in ``Gold'', in such condition subwords in a minimal span whose edges are aligned to ones in the ``Gold'' are regard as a virtual word, and tagged with the additional labels.

\textbf{Word segmentation decoder:} After getting the tagged sequence, we adopt a simple segmentation strategy based on label 'E', '-E' and 'S': while scanning the label sequence, we output a word by concatenating subwords in buffer when encounter label in 'E, -E, S' or store the subword in buffer. It is shown that the decoding rule is easy for implementation and  roboust to tagging errors of the Refiner.

We concatenate embeddings of subwords with its feature embedding to form inputs of Neural network sequence tagger, see bottom side of Figure \ref{fig:refiner}.

\section{Pipeline Description}
We introduce the pipeline of building the translation systems from data preprocessing to system trainning in this section.

\subsection{Data Preprocessing}
For Tibetan word segmentation task, MLWS 2017 provides a 1w sentences training Corpus.
We used this corpora to train our segmentation systems combined with 73w Tibetan sentences for machine translation and 2w Tibetan sentences extracted from Part of speech tagging corpus in-house.
For baseline system, we used our in-house word segmentor called ``tiSegmenter'' to do the word segmentation. 
To train BPE model for Refiner's subword Preprocessing unit, we used 1664675 monolingual sentences clawed from tibetan websites.
We filter the sentences of poor quality and the sentences that are too long (more than 120 words) . Before the training of neural network, we transformed these sentences into subword sequences by the pipeline of baseline system and BPE model, 
and generate the corresponding label sequences for supervised training.

\subsection{System configuration}
We set the vocabulary size for BPE as 2w, 1w, 0.5w, 0.1w respectively for experiment, and find that 2w is the best for Refiner's performance, and set is as default bellow.
For our system is serially trained on the single GPU with restricted memory space, and the vocabulary size is set to 18559. The words that out of the vocabulary are represented by the ``UNK'' symbol. We stack 8 modified LSTM units as the elements of neural network, cells with the same unrolling direction share parameters, hidden size of lstm is 512, word and feature embedding size are set to 256 to enable residual operation in the first layer of neural network.

\subsection{System evaluation}
We took the evaluation scripts \footnote{http://sighan.cs.uchicago.edu/bakeoff2005/} used in the 2nd International Chinese Word Segmentation Bakeoff to calculate the F-score of segmentation results. We splited the training set into two parts, 1k validation set for model selection during training while the remaining was for system training.

\subsection{Training Details}
The sentence length for training systems is up to 120, for long sentence exceeding such limitation we splited it into shorter ones by special terminator tokens.
Refiner's neural network classifier is trained by the open source toolkit mxnet \footnote{https://mxnet.incubator.apache.org/}
We initialized the weight in lstm unit with Xavier scheme of factor type "in" and magnitude 2.34, and initialized word embedding matrix by embeddings learned from preprocessed monolingual corpus using fasttext \footnote{https://github.com/facebookresearch/fastText}.
Parameters are updated by Mini-batch Gradient Descent and the learning rate is controlled by the AdaDelta algorithm \cite{Zeiler2012ADADELTA} with the decay constant $\rho=0.9$ and the denominator constant $\epsilon=1e-5$. 
The batch size is $150$.
Dropout strategy \cite{srivastava2014dropout} is applied to the input of lstm unit with the dropout rate 0.1 to avoid over-fitting.
The gradients of parameters were scaled by rate 0.1 and clipped to range [-1.0, 1.0] to avoid gradients explosion.

\section{Experimental Results}
\subsection{Baseline system}
For the time-consuming nature of training baseline system, we reuse the model in-house without extra training procedure.
We test the baseline system on the validation set.
The performance of the system on the set is presented in Table \ref{table:baseline}.
\begin{table}[htb]
\begin{center}
\begin{tabular}{l|c}
\hline
\textbf{Data} &\textbf{Validation Set} \\
\hline
\textbf{F-score } &89.9\\
\hline
\end{tabular}
\end{center}
\caption{\label{table:baseline} The baseline performance on the validation set}
\end{table}

We took part in MLWS2017  tibetan word segmentation shared task held by the 16th national conference on ethnic minority language information    processing of china， our baseline system's performance on the test set on news domain provided by  MLWS2017 is presented in Table \ref{table:baseline1}.

\begin{table}[!htb]
\begin{center}
\begin{tabular}{l lc}
\hline
\textbf{Data} &\textbf{Validation set} \\
\hline
\textbf{F-score  } &90.08\\
\cline{2-2} 
\textbf{Precision  } &90.32 \\
\cline{2-2} 
\textbf{Recall  } &89.83 \\
\hline
\end{tabular}
\end{center}
\caption{\label{table:baseline1} The baseline performances on the test set .}
\end{table}

\subsection{Baseline system + Refiner}
Based on the baseline system, we build the training corpus for Refiner.
After the training of BPE model we generate training data for the neural network classifier.
It takes about 1 hour to train the classifier on a single NVIDIA GeForce GPU. Our model reaches its best performance on validation set at epoch 6.

We test the system proposed in the paper by introducing a Refiner after the baseline system on the validation set and chose the best one.
Table \ref{table:refiner} presents the performance of best system on the validation set.
It is shown that our approach can enhance the system performance over baseline model.
The Refiner module improves the baseline system by +6.6 F-score points on validation set.

\begin{table}[!htb]
\begin{center}
\begin{tabular}{ll}
\hline
\textbf{Data} &\textbf{Test set} \\
\hline
\textbf{F-score} &96.5$^{+6.6}$\\
\cline{2-2} 
\textbf{OOV Rate} &0.811 \\
\cline{2-2} 
\textbf{OOV Recall Rate} &0.969 \\
\cline{2-2} 
\textbf{IV Recall Rate} &0.980 \\
\hline
\end{tabular}
\end{center}
\caption{\label{table:refiner} The model performances on the validation set .}
\end{table}

The system's performance on the test set  provided by  MLWS2017 is presented in Table \ref{table:refiner1}, the Refiner module improves the baseline system by +2.48 F-score points on test set. Our system was superior to almost all other systems to get second place in MLWS2017 shared task.
 
\begin{table}[!htb]
\begin{center}
\begin{tabular}{l l}
\hline
\textbf{Data} &\textbf{Test set} \\
\hline
\textbf{F-score  } &92.56$^{+2.48}$\\
\cline{2-2} 
\textbf{Precision  } &93.32 \\
\cline{2-2} 
\textbf{Recall  } &91.32 \\
\hline
\end{tabular}
\end{center}
\caption{\label{table:refiner1} The model performances on the test set .}
\end{table}

\section{Conclusion}
We present a CASICT tibetan word segmentation system for the MLWS2017  tibetan word segmentation shared task .
A baseline tibetan segmentation system use a Refiner module to imporve its performance further in a simple and efficient way.
A argumented label set is proposed a deal with the error propogation form baseline system as well as a simple strategy for robust decoding.
A BPE algorithm based subword preprocessing unit is proposed for the Refiner.
A LSTM unit with a gated residual connection is employed to build the deep neural network for subword sequence labeling. Experiment shows that the Refiner can learn to fix segmentation errors of baseline system and result in a significant performance gain.

\section*{Acknowledgments}
\noindent
Qun Liu's work is partially supported by Science Foundation Ireland in the ADAPT Centre for Digital Content Technology (www.adaptcentre.ie) at Dublin City University funded under the SFI Research Centres Programme (Grant 13/RC/2106) co-funded under the European Regional Development Fund.

\bibliography{llncs}
\bibliographystyle{splncs03}

\end{document}